# Positive Real Synthesis of Networked Control System : An LMI Approach

**Bambang Riyanto\*** and **Imam Arifin\***

\*School of Electrical Engineering and Informatics
Institut Teknologi Bandung, Bandung, INDONESIA.
e-mail: briyanto@lskk.ee.itb.ac.id , arifin-i@ee.its.ac.id

**Abstract**

This paper presents the positive real analysis and synthesis for Networked Control Systems (NCS) in discrete-time. Based on the definition of passivity, the sufficient condition of NCS is given by stochastic Lyapunov functional. The controller via state feedback is designed to guarantee the stability of NCS and closed-loop positive realness. It is shown that a mode-dependent positive real controller exists if a set of coupled linear matrix inequalities has solutions. The controller can be then constructed in terms of the solutions.

## 1 Introduction

The recent advancement of computing, communication and sensing has spurred the development of Networked Control Systems (NCS). NCS are spatially distributed systems in which communication between plants, sensors, actuators and controllers occurs through a shared band-limited digital communication networks [3,11]. These networks are usually shared by a number of feedback loops. In traditional control systems, these connections are established via point-to-point cable wiring. Compared with the point-to-point wiring, the introduction of communication networks has many advantages, such as high system testability and resource utilization, as well as low weight, space, power, and wiring requirements and easy system diagnosis and maintenance. The increasingly fast convergence of sensing, computing and (wireless) communication on cost effective, low power, small-size devices, is also quickly enabling a surge of new control applications. Consequently NCS has finding application in a broad range of areas such as mobile sensor networks, remote surgery, multiple mobile autonomous robotics, multiple unmanned aerial vehicles(UAV) formation, large scale distributed industrial processes and automation, computer integrated manufacturing systems, tele-operation and tele-control, intelligent vehicle systems, sattelite clusters, etc[8]. In an NCS, it also makes it possible to distribute processing functions and computing loads into several small units. Recently, modeling, analysis and control of networked control systems with limited communication capability has emerged as a topic of significant interest to control community. There are a number of key issues that make control over communication networks distinct from traditional control systems : limited packet rate, sampling and network delay, packet dropout and system architecture. The traditional point-to-point feedback control theory should be extended for feedback control system for which connections among sensors, actuators, and controllers are shared through band-limited communication networks. With the advent of cheap, small, low-power processors with communication capabilities, it has become possible to endow sensors and actuators with processing power and ability to communicate with remote controllers through shared/multi-purpose networks [9].

In this paper we propose a new control design method for networked control system which satisfies positive-real (passivity) requirements, in addition to stability. Important aspects of networked control systems, namely random delays and packet losses, are taken into consideration in the passive control design. The networked control system is modeled using Markov Jump Linear System, where packet loss is modeled as two-state **Bernoulli** process [3].

The concept of passivity or positive realness plays an important role in system, signal and control theory, and arises naturally in many areas of science and engineering [12]. The idea originates in traditional circuit theory. In general, positive realness corresponds to passivity for linear time-invariant dynamical systems. The main motivation for studying passivity comes from robust, adaptive and nonlinear control problems. Parallel to small gain theorem, passivity of certain closed-loop transfer functions will ensure the overall stability of feedback systems [1]. Therefore, if a certain closed-loop transfer function matrix can be made passive by using some controllers, then the closed-loop system will be guaranteed to be robust with respect to all passive uncertainties. The passivity requirements are also important in a number of practical cases where the control designer has information about phase uncertainties.

The synthesis problem considered in this paper is how to construct a feedback controller for a networked control system such that the resulting closed-loop system satisfies passivity. Based on the definition of passivity under state-space representation of jump linear systems, the passivity conditions are firstly obtained using stochastic Lyapunov functional technique. The conditions also show that the passivity of jump systems guarantees the stability. A kind of mode-dependent controllers suited for a networked control system is proposed via state feedback to ensure the passivity of the resulting closed-loop systems. By manipulation of matrix and change of controller variables, the existence conditions and the synthesis methods for the mode-dependent controllers can be formulated in terms of a set of solutions of coupled linear matrix inequalities, which can be effectively solved using available LMI-software.

The remainder of the paper proceeds as follows : In Section 2, networked control analysis and synthesis problems are formulated. In Section 3, a new approach for stabilization is presented for networked control system to ensure closed-loop passivity. Finally, we conclude the result of this paper in Section 4.

## 2  Networked Control System Problem Statement and Analysis

The problem is formulated as a remote control system. Formulation of the networked control system as a Markov jump linear systems follows from [2,3]. The plant is remotely controlled by the controller connected via a shared communication channel. It is assumed that there are multiple sensors and actuators communicating with the controller; however, as a result of the sequential nature of the channel, only one of them can transmit a message at any discrete-time instant. To meet this requirement efficiently, we employ a periodic communication scheme[3].

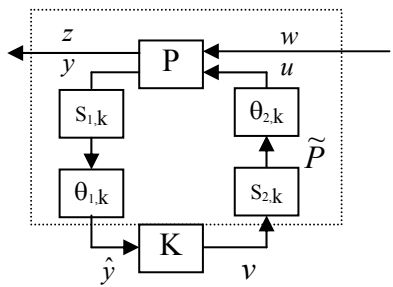

Figure 1. Remote Control System [3]

Generalized plant P is described in the form of discrete-time system :

$$\begin{aligned} x(k+1) &= Ax(k) + B_1 w(k) + B_2 u(k) \\ z(k) &= C_1 x(k) + D_{11} w(k) + D_{12} u(k) \\ y(k) &= x(k) \end{aligned} \quad (1)$$

where $x(k) \in \Re^n$ is state, $w(k) \in \Re^{m_1}$ is exogenous input, $u(k) \in \Re^{m_2}$ is control input, $z(k) \in \Re^{p_1}$ is controlled output and $y(k) \in \Re^{p_2}$ is measurement output

The plant P is remotely controlled by the controller connected via a shared communication channel. It is assumed that there are multiple sensors and actuators communicating with the controller, however, as a result of the sequential nature of channel, only one of them can transmit a message at any discrete time instant. To meet this requirement efficiently, we employ a periodic communication scheme.

Suppose there are $p_2$ sensors and $m_2$ actuators, then the period be $N \geq m_2 + p_2$. If each sensor is capable to transmit messages, then the *switching pattern* for the sensor side defined by a vector $s_1 \in I_{p_2+1}^N$. This specifies that at time $k$, the sensor indexed as $s_1(\mod(k,N)+1)$ is allowed to send a message. If $s_1(\mod(k,N)+1)$ is zero, it means there is no communication takes place. For example, let $N = 4$, $p_2 = 2$, and $s_1 = [2, 1, 0, 2]$. In this case, sensor 1 transmits at $k = 1, 5, \ldots$ while sensor 2 transmits at $k = 0, 3, \ldots$, and there is no communication at $k = 2, 6, \ldots$.

Similarly, the switching pattern $s_2 \in I_{m_2+1}^N$ for the $m_2$ actuators that determines the periodic transmission from the controller to the actuators with the same period $N$. We assume that, at any given time, at most only one message can be transmitted by the sensors or the controller. In terms of the switching patterns, this means that $s_i(k) \neq 0$ iff $s_j(k) = 0$ for $i \neq j$ and all $k$.

We now give some notation for the periodic switchings. Define two sets $\{s_{1i}\}$ and $\{s_{2i}\}$ of vectors given as follows:

$$s_{1i} = [0...010...0] \in \Re^{1 \times p_2} \quad ; i = 1,...,p_2$$
$$s_{2j} = [0...010...0]^T \in \Re^{m_2 \times 1} \quad ; j = 1,...,m_2$$

where $s_{1i}$ and $s_{2j}$ have the $i^{th}$ and $j^{th}$ elements equal to 1, respectively, and the rest are zero.
The matrices $S_{1,k}$ and $S_{2,k}$ in Figure 1 are defined as :

$$S_{1,k} := s_{1i} \quad ; s_1(\mod(k,N)+1) = i$$
$$S_{2,k} := s_{2j} \quad ; s_2(\mod(k,N)+1) = j$$

for $k \in Z_+$. We refer to them as the switches. Note that these are $N$-periodic matrices.

Another feature of the channel is that it is lossy: Transmitted messages randomly are lost due to congestion or delay. Let $\theta_{1,k}, \theta_{2,k} \in \{0, 1\}$ be the stochastic processes that represent the losses, from the sensors to the controller and from the controller to the actuators. If $\theta_{i,k} = 0$, then the message at time $k$ is lost; otherwise, it arrives. We assume that these are independent and identically distributed (i.i.d.) **Bernoulli** processes specified by

$$\alpha_1 := \text{Prob}\{\theta_{1,k} = 0\}$$
$$\alpha_2 := \text{Prob}\{\theta_{2,k} = 0\} \quad ; k \in Z_+$$

The overall plant including the switches $S_1$ and $S_2$ and the message loss processes $\theta_1$ and $\theta_2$ is periodically time varying with period $N$ and with random switchings.

The state space equation of $\tilde{P}$ can be expressed as:
$$\begin{aligned} x(k+1) &= Ax(k) + B_1 w(k) + \theta_{2,k} B_2 S_{2,k} v(k) \\ z(k) &= C_1 x(k) + D_{11} w(k) + \theta_{2,k} D_{12} S_{2,k} v(k) \\ \hat{y}(k) &= \theta_{1,k} S_{1,k} x(k) \end{aligned} \quad (2)$$

In Figure 1, $K$ is controller to be designed, with $N$-period, receives acknowledgement from actuator based on control input $u_k$.
The controller takes the form as follows:
$$v(k) = K \hat{y}(k) \quad (3)$$

In this paper, we employ the following notion of stability.
Definition 1 [2] : For the system given by
$$\begin{bmatrix} x(k+1) \\ z(k) \end{bmatrix} = \begin{bmatrix} A_{\theta(k)} & B_{\theta(k)} \\ C_{\theta(k)} & D_{\theta(k)} \end{bmatrix} \begin{bmatrix} x(k) \\ w(k) \end{bmatrix} \quad (4)$$
with $w \equiv 0$, the equilibrium point at $x=0$ is
1. mean square stable if for every initial state $(x_0, \theta_0)$
$$\lim_{k \to \infty} E\left[\|x(k)\|^2 \big| x_0, \theta_0\right] = 0$$
2. stochastically stable if for every initial state $(x_0, \theta_0)$,
$$\sum_{k=0}^{\infty} E\left[\|x(k)\|^2 \big| x_0, \theta_0\right] < \infty.$$ In other words, $\|x\|_2 < \infty$ for every initial state.
3. exponentially mean square stable if for every initial state $(x_0, \theta_0)$, there exists constants $0 < \alpha < 1$ and $\beta > 0$ such that $\forall k \geq 0$, $E\left[\|x(k)\|^2 \big| x_0, \theta_0\right] < \beta \alpha^k \|x_0\|^2$
4. almost surely stable if for every initial state $(x_0, \theta_0)$,
$$\Pr\left[\lim_{k \to \infty} \|x(k)\| = 0\right] = 1$$

It is known that the first three definitions of stability are actually equivalent for an MJLS. We refer to the equivalent notions of mean square, stochastic, and exponential mean square stability as second moment stability. Moreover, Second Moment Stability (SMS) is sufficient but not necessary for almost sure stability. In the remainder of this paper, references to stability will be in the sense of second moment stability. The major motivation for this choice is that straightforward necessary and sufficient conditions, given later, exists to check for SMS but not for almost sure stability [2].

*Lemma 1 [3]* : For the system (4), the origin is stochastically stable if and only if there exists an $N$-periodic matrix $P_k \in R^{nxn}$ such that $P_k = P_k^T > 0$ and
$$\sum_{i \in I_M} \alpha_i A_{k,i}^T P_{k+1} A_{k,i} - P_k < 0 \quad ; k \in I_N$$

where $I_N := \{0,1,...,N-1\}$
The theorem states that SMS is equivalent to finding $N$ positive definite matrices which satisfy $N$ coupled, discrete Lyapunov equations. It is interesting to note that stability of each mode is neither necessary nor sufficient for the system to be SMS. The condition simplifies under an additional assumption on the Markov process.

Definition 2 [1] : The Markovian jump linear system (4) with $u(k)=0$, is said to be strictly passive, if every $T>0$, with zero initial condition $x_0=0$, it satisfies
$$E\left\{\sum_{k=0}^{T} w(k)^T z(k)\right\} > 0 \quad (5)$$

Furthermore, it is said to be strictly passive with dissipation $\eta$ if
$$E\left\{\sum_{k=0}^{T} (w(k)^T z(k) - \eta w(k)^T w(k))\right\} > 0 \quad (6)$$

*Remark 1*. If the notation '>' is replaced by '≥', the above strictly passive definition is referred to as passive definition. This paper focuses on strictly passive problems, and often the strictly passivity is referred to as passivity wherever no confusion arises.

Following analysis result is related to passivity of Markovian jump systems.

*Asumption 1*. $D_{wi} + D_{wi}^T > 0$ ; $i \in T$

*Lemma 2*. For $\eta \geq 0$, the markovian jump linear systems (4) is said to be strictly passive with dissipation $\eta$ if, there exists a set of positive definite symmetric matrices, $P_i, i \in T$ such that
$$\Theta_i = \begin{bmatrix} A_i^T \overline{P}_i A_i - P_i & A_i^T \overline{P} B_{wi} - C_i^T \\ B_{wi}^T \overline{P}_i A_i - C_i & B_{wi}^T \overline{P}_i B_{wi} + 2\eta I - D_{wi}^T - D_{wi} \end{bmatrix} < 0 \quad (7)$$

*Proof*. Based on Lemma 1, the Markovian jump system (4) is stochastically stable if, there exists a set of positive definite symmetric matrices $P_i, i \in T$ such that
$$A_i^T \overline{P}_i A_i - P_i < 0 \quad (8)$$
with
$$\overline{P}_i = \sum_{j=1}^{N} \pi_{ij} P_j$$

For given positive definite symmetric matrices $P_i \in R^{nxn}$, set up the stochastic Lyapunov functional as
$$V(k)(\theta(k), r(k)) = x^T(k) P(\theta(k)) x(k) \quad (9)$$
While $\theta(k) = i$, denoting $P(\theta(k))$ as $P_i$, considering Markov jump system (4), and letting $\theta(k+1) = j$, we have
$$\begin{aligned} &E\{V(k+1)(x(k+1), \theta(k+1)) | x(k), \theta(k)\} - V(k)(x(k), \theta(k)) \\ &= \sum_{j=1}^{N} p_{ij} x^T(k+1) P_j x(k+1) - x^T(k) P_i x(k) \\ &= x^T(k)(A_i^T \overline{P}_i A_i - P_i) x(k) + x^T(k) A_i^T \overline{P}_i B_{wi} w(k) \\ &\quad + w^T(k) B_{wi}^T \overline{P}_i A_i x(k) + w^T(k) B_{wi}^T \overline{P}_i B_{wi} w(k) \end{aligned} \quad (10)$$

Following condition is constructed to satisfy the Definition 2 above.

$$E\{\sum_{k=0}^{T}[V(k+1)(x(k+1),\theta(k+1))|x(k),\theta(k)] - V(k)(x(k),\theta(k)) - z^T(k)w(k)$$
$$- w^T(k)z(k) + 2\eta w^T(k)w(k)\}$$
$$= E\{\sum_{k=0}^{T}[x^T(k)(A_i^T\overline{P}_iA_i - P_i)x(k) + x^T(k)A_i^T\overline{P}_iB_{wi}w(k) + w^T(k)B_{wi}^T\overline{P}_iA_ix(k)$$
$$+ w^T(k)B_{wi}^T\overline{P}_iB_{wi}w(k)] - x^T(k)C_i^Tw(k) - w^T(k)C_ix(k)$$
$$+ w^T(k)(2\eta I - D_{wi}^T - D_{wi})w(k)\}$$
$$= \sum_{k=0}^{T}\zeta^T(k)\Theta_i\zeta(k) < 0$$

(11)

where $\zeta(k) = [x^T(k) \quad w^T(k)]^T$, obviously, condition (11) holds if $\Theta_i < 0$. This completes the proof..

□

*Lemma 3* [7]. The block matrix $\begin{bmatrix} P & M \\ M^T & Q \end{bmatrix}$ is negative definite if and only if

$$Q < 0$$
$$P - MQ^{-1}M^T < 0$$

(12)

In the sequel, $P - MQ^{-1}M^T$ will be referred to as the Schur complement.

### 3  Positive Real Synthesis of Networked Control System

In this section, we present the solution to the remote control synthesis problem.

For the system (2) and controller (3), we have the state space equation for the closed loop system $F_l(\widetilde{P},K)$

$$\overline{x}(k+1) = \overline{A}_{k,\theta_1(k),\theta_2(k)}\overline{x}(k) + \overline{B}_{k,\theta_1(k),\theta_2(k)}w(k)$$
$$z(k) = \overline{C}_{k,\theta_1(k),\theta_2(k)}\overline{x}(k) + \overline{D}_{k,\theta_1(k),\theta_2(k)}w(k)$$

(13)

where $\overline{x}_k \in R^{2n}$ is state given by $\overline{x}_k := [x_k^T \quad \hat{x}_k^T]^T$ and ;

$$\overline{A}_{k,\theta_1(k),\theta_2(k)} := [A + \theta_{2,k}B_2S_{2,k}K\theta_{1,k}S_{1,k}]$$
$$\overline{B}_{k,\theta_1(k),\theta_2(k)} := B_1$$
$$\overline{C}_{k,\theta_1(k),\theta_2(k)} := [C_1 + \theta_{2,k}D_{12}S_{2,k}K\theta_{1,k}S_{1,k}]$$
$$\overline{D}_{k,\theta_1(k),\theta_2(k)} := D_{11}$$

Notice that A and C system matrices depend on both $\theta_1$ and $\theta_2$.

As mentioned earlier, this system consists of 4 modes determined by the original process pairs $(\theta_{1,k},\theta_{2,k})$. Let $\theta_k := (\theta_{1,k},\theta_{2,k})$, which is associated with the probability $\alpha_{i,j} := \Pr ob\{\theta_k = (i,j)\}$, $i,j \in \{0,1\}$, where

$$\alpha_{0,0} = \alpha_1\alpha_2 \qquad \alpha_{1,0} = (1-\alpha_1)\alpha_2$$
$$\alpha_{0,1} = \alpha_1(1-\alpha_2) \qquad \alpha_{1,1} = (1-\alpha_1)(1-\alpha_2)$$

(14)

We can apply Lemma 2 to this system to obtain the synthesis result.

The following is the main result of the paper and provides a solution to the positive real remote control problem.

*Theorem 1*. For closed loop system (13) and dissipation $\eta \geq 0$, there exists a set of $P_i$ satisfying condition $\Theta_i < 0$, if and only if there exists a set of positive definite symmetric matrices $X_i \in R^{n \times n}$, matrices $Y_i \in R^{m \times n}$ satisfying

$$\begin{bmatrix} -X & (.)^T & (.)^T & (.)^T & (.)^T & (.)^T \\ -[C_1X + \hat{D}_{12}Y] & 2\eta I - (D_{11})^T - D_{11} & (.)^T & (.)^T & (.)^T & (.)^T \\ \sqrt{\alpha_{00}}[AX + \hat{B}_2Y] & B_1 & -X & 0 & 0 & 0 \\ \sqrt{\alpha_{01}}[AX + \hat{B}_2Y] & B_1 & 0 & -X & 0 & 0 \\ \sqrt{\alpha_{10}}[AX + \hat{B}_2Y] & B_1 & 0 & 0 & -X & 0 \\ \sqrt{\alpha_{11}}[AX + \hat{B}_2Y] & B_1 & 0 & 0 & 0 & -X \end{bmatrix} < 0$$

(15)

By manipulation of matrix and change of controller variables, the existence conditions and the synthesis methods for the mode-dependent controllers can be formatted in terms of a set of solutions of coupled linear matrix inequalities, which can be effectively solved using LMI-software. Furthermore, the mode-dependent state feedback passive controller is given by $v_k = K\hat{y}_k$, where $K = Y(\theta_{1,k}S_{1,k}X)^{-1}$.

*Proof*. Suppose there exists a set of positive definite symmetric matrices $P_i$, $i \in T$, satisfying condition $\Theta_i < 0$ in equality (7),i.e.

$$\begin{bmatrix} A^T\overline{P}A - P & A^T\overline{P}B - C^T \\ B^T\overline{P}A - C & B^T\overline{P}B + 2\eta I - D^T - D \end{bmatrix} < 0$$

(16)

$$\overline{P} = \sum_{i,j=0}^{1}\alpha_{ij}P$$

Applying Schur complements, above inequality (16) is equivalent to the following inequality

$$\begin{bmatrix} -P & (.)^T & (.)^T & (.)^T & (.)^T & (.)^T \\ -\overline{C}_{k,\theta_1(k),\theta_2(k)} & 2\eta I - (\overline{D}_{k,\theta_1(k),\theta_2(k)})^T - \overline{D}_{k,\theta_1(k),\theta_2(k)} & (.)^T & (.)^T & (.)^T & (.)^T \\ \sqrt{\alpha_{00}}\overline{A}_{k,\theta_1(k),\theta_2(k)} & \overline{B}_{k,\theta_1(k),\theta_2(k)} & -P^{-1} & 0 & 0 & 0 \\ \sqrt{\alpha_{01}}\overline{A}_{k,\theta_1(k),\theta_2(k)} & \overline{B}_{k,\theta_1(k),\theta_2(k)} & 0 & -P^{-1} & 0 & 0 \\ \sqrt{\alpha_{10}}\overline{A}_{k,\theta_1(k),\theta_2(k)} & \overline{B}_{k,\theta_1(k),\theta_2(k)} & 0 & 0 & -P^{-1} & 0 \\ \sqrt{\alpha_{11}}\overline{A}_{k,\theta_1(k),\theta_2(k)} & \overline{B}_{k,\theta_1(k),\theta_2(k)} & 0 & 0 & 0 & -P^{-1} \end{bmatrix} < 0$$

For matrices $\overline{A}_{k,\theta_1(k),\theta_2(k)}, \overline{B}_{k,\theta_1(k),\theta_2(k)}, \overline{C}_{k,\theta_1(k),\theta_2(k)}, \overline{D}_{k,\theta_1(k),\theta_2(k)}$ in (13) , we obtain

$$\begin{bmatrix} -P & (.)^T & (.)^T & (.)^T & (.)^T & (.)^T \\ -[C_1 + \theta_{2,k}D_{12}S_{2,k}K\theta_{1,k}S_{1,k}] & 2\eta I - (D_{11})^T - D_{11} & (.)^T & (.)^T & (.)^T & (.)^T \\ \sqrt{\alpha_{00}}[A + \theta_{2,k}B_2S_{2,k}K\theta_{1,k}S_{1,k}] & B_1 & -P^{-1} & 0 & 0 & 0 \\ \sqrt{\alpha_{01}}[A + \theta_{2,k}B_2S_{2,k}K\theta_{1,k}S_{1,k}] & B_1 & 0 & -P^{-1} & 0 & 0 \\ \sqrt{\alpha_{10}}[A + \theta_{2,k}B_2S_{2,k}K\theta_{1,k}S_{1,k}] & B_1 & 0 & 0 & -P^{-1} & 0 \\ \sqrt{\alpha_{11}}[A + \theta_{2,k}B_2S_{2,k}K\theta_{1,k}S_{1,k}] & B_1 & 0 & 0 & 0 & -P^{-1} \end{bmatrix} < 0$$

Define $\hat{K} = K\theta_{1,k}S_{1,k}$ and $\hat{B} = \theta_{2,k}B_2 S_{2,k}$, then the inequality becomes

$$\begin{bmatrix} -P & (.)^T & (.)^T & (.)^T & (.)^T & (.)^T \\ -[C_1 + \hat{D}_{12}\hat{K}] & 2\eta I - (D_{11})^T - D_{11} & (.)^T & (.)^T & (.)^T & (.)^T \\ \sqrt{\alpha_{00}}[A + \hat{B}_2\hat{K}] & B_1 & -P^{-1} & 0 & 0 & 0 \\ \sqrt{\alpha_{01}}[A + \hat{B}_2\hat{K}] & B_1 & 0 & -P^{-1} & 0 & 0 \\ \sqrt{\alpha_{10}}[A + \hat{B}_2\hat{K}] & B_1 & 0 & 0 & -P^{-1} & 0 \\ \sqrt{\alpha_{11}}[A + \hat{B}_2\hat{K}] & B_1 & 0 & 0 & 0 & -P^{-1} \end{bmatrix} < 0$$

Pre- and post-multiplying the inequality by block-diagonal matrix $diag\{P^{-1}, I, I, I, I, I\}$, letting $X = P^{-1}$, $Y = KX$, the coupled linear matrix inequalities (15) are obtained.

□

Based on Theorem 1, the controller is constructed as follows. First, solve LMI (15) for positive definite symmetric matrices $X_i$ and $Y_i$ using available LMI solver. Then, using these solutions the controller is obtained via $K = Y(\theta_{1,k}S_{1,k}X)^{-1}$.

## 4 Conclusions

In this paper we have presented a new approach to analysis and synthesis problems for networked control systems in discrete-time. The networked control system is modeled using Markov Jump Linear System. We consider state feedback controller based on positive real synthesis. The sufficient condition on positive real of the system presented by stochastic Lyapunov functional. The positive real controller via state feedback, which guarantees the stability, obtained by using the solutions of a set of coupled linear matrix inequality.